\documentclass[conference]{IEEEtran}
\IEEEoverridecommandlockouts

\usepackage{cite}
\usepackage{amsmath,amssymb,amsfonts}
\usepackage{algorithmic}
\usepackage{multirow,array,enumitem,adjustbox}
\usepackage{booktabs,threeparttable,makecell}
\usepackage{graphicx}
\usepackage{textcomp}
\usepackage[table,xcdraw]{xcolor}
\usepackage{cleveref}

\def\BibTeX{{\rm B\kern-.05em{\sc i\kern-.025em b}\kern-.08em
    T\kern-.1667em\lower.7ex\hbox{E}\kern-.125emX}}

\makeatletter
\newcommand{\linebreakand}{%
  \end{@IEEEauthorhalign}
  \hfill\mbox{}\par
  \mbox{}\hfill\begin{@IEEEauthorhalign}
}
\makeatother

\begin{document}

\title{A Multi-scenario Attention-based Generative Model for Personalized Blood Pressure Time Series Forecasting\\
\thanks{$^\ast$ indicates the corresponding author: Dan Wu, Tel: 86-755-86392207; e-mail: dan.wu@siat.ac.cn. \\   \null\quad\ \ This work was supported by the National Key Research and Development Program (2021YFF0703704); Guangdong Basic and Applied Basic Research Foundation (2022A1515011217); National Natural Science Foundation
of China (U2241210 and U1913210).}
}

\author{\IEEEauthorblockN{1\textsuperscript{st} Cheng Wan}
\IEEEauthorblockA{\small \textit{Shenzhen Institute of Advanced Technology} \\
\textit{Chinese Academy of Science} \\
Shenzhen, China \\
\textit{Georgia Institute of Technology}\\
Atlanta, USA\\
jouiney666@gmail.com}
\and
\IEEEauthorblockN{2\textsuperscript{nd} Chenjie Xie}
\IEEEauthorblockA{\small \textit{Shenzhen Institute of Advanced Technology} \\
\textit{Chinese Academy of Science} \\
Shenzhen, China \\
\textit{Southern University of Science and Technology}\\
Shenzhen, China\\
cj.xie@siat.ac.cn}
\and
\IEEEauthorblockN{3\textsuperscript{rd} Longfei Liu}
\IEEEauthorblockA{\small \textit{Shenzhen Institute of Advanced Technology } \\
\textit{Chinese Academy of Science} \\
Shenzhen, China \\
\textit{University of Chinese Academy of Sciences}\\
Beijing, China\\
lf.liu@siat.ac.cn}
\linebreakand
\IEEEauthorblockN{4\textsuperscript{th} Dan Wu$^{\ast}$}
\IEEEauthorblockA{\small \textit{Shenzhen Institute of Advanced Technology } \\
\textit{Chinese Academy of Science} \\
Shenzhen, China \\
dan.wu@siat.ac.cn}
\and
\IEEEauthorblockN{5\textsuperscript{th} Ye Li}
\IEEEauthorblockA{\small \textit{Shenzhen Institute of Advanced Technology } \\
\textit{Chinese Academy of Science} \\
Shenzhen, China \\
ye.li@siat.ac.cn}
}
\maketitle
\begin{abstract}
Continuous blood pressure (BP) monitoring is essential for timely diagnosis and intervention in critical care settings. However, BP varies significantly across individuals, this inter-patient variability motivates the development of personalized models tailored to each patient's physiology. In this work, we propose a personalized BP forecasting model mainly using electrocardiogram (ECG) and photoplethysmogram (PPG) signals. This time-series model incorporates 2D representation learning to capture complex physiological relationships. Experiments are conducted on datasets collected from three diverse scenarios with BP measurements from 60 subjects total. Results demonstrate that the model achieves accurate and robust BP forecasts across scenarios within the Association for the Advancement of Medical Instrumentation (AAMI) standard criteria. This reliable early detection of abnormal fluctuations in BP is crucial for at-risk patients undergoing surgery or intensive care. The proposed model provides a valuable addition for continuous BP tracking to reduce mortality and improve prognosis.
\end{abstract}


\begin{figure*}[t]
\centering
\includegraphics[width=0.9\linewidth,scale=1.0]{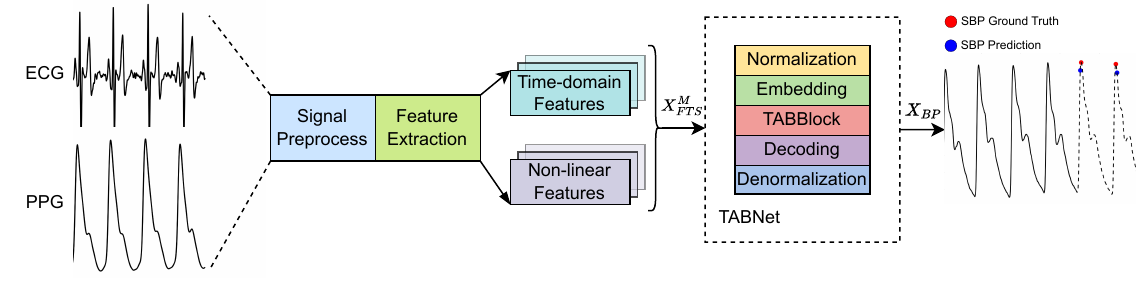}
\caption{\textbf{Illustration of the proposed BP forecasting framework.} $X_{\text{FTS}}^{M}$ represents the 38 extracted features from time and non-linear domains. The red dots on the true BP waveform drawn by the wavy line represents the ground truth of BP in the incoming two cycles while the blue dots represents the prediction BP results from models.}
\label{fig:totalframe}
\end{figure*}
\section{Introduction}
\label{sec:intro}
Continuous monitoring of blood pressure (BP) is crucial in critical care settings such as operating rooms and intensive care units~\cite{pinsky2003hemodynamic}. Sudden BP changes can signal patient deterioration or adverse treatment responses~\cite{araghi2006arterial}. However, intermittent cuff-based measurements may miss acute changes, driving research into cuffless BP estimation using physiological signals~\cite{mukkamala2015toward}. Generalized models have been developed but often fail to consider individual variations in BP regulation~\cite{wang2014cuff}, making it difficult to achieve consistent performance in real-world applications. For instance, Slapni{\v{c}}ar et al.~\cite{slapnivcar2019blood} proposed a PPG-based method with a mean absolute error (MAE) of 9.43 mmHg for systolic blood pressure (SBP), and Sun et al.~\cite{sun2023systolic} reported a standard deviation (SD) of 7.97 mmHg for SBP, highlighting the limitations in accuracy of generalized BP models. To address this, personalized BP estimation has gained increasing attention~\cite{leitner2021personalized, chiang2021using}. Leitner et al.~\cite{leitner2021personalized} developed a BP-CRNN-Transfer model using pre-training and fine-tuning, which improved prediction accuracy but involved complex steps unsuitable for clinical practice. Wang et al.~\cite{wang2023personalized} designed a lightweight personalized model for long-term BP tracking, though it was limited to a single scenario. Pediaditis et al.~\cite{pediaditis2021cuff} demonstrated that individualized models significantly reduced errors compared to group-level models.

In this work, we propose a personalized BP forecasting model with high accuracy across diverse scenarios, utilizing electrocardiogram (ECG) and photoplethysmogram (PPG) signals. ECG provides cardiac activity data, while PPG captures peripheral vascular dynamics~\cite{slapnivcar2019blood}. Combining these signals improves BP estimation accuracy~\cite{dash2009automatic, wu2018continuous}. We frame BP forecasting as a time series modeling problem, aiming to predict future BP values based on past measurements. Accurate short-term forecasting enables early detection of critical changes in at-risk patients~\cite{clifford2015physionet}, allowing timely clinical intervention.

The main contributions of this work are:
\begin{itemize}
    \item We propose a BP prediction generative model effective in multiple medical scenarios, enabling non-invasive assessment and flexible prediction timing for earlier clinical intervention.
    \item We introduce multi-domain feature fusion, incorporating time-domain and nonlinear features, providing richer information for improved BP forecasting accuracy.
    \item We develop a personalized BP prediction model with an attention mechanism, allowing the model to focus on distinctive individual features for accurate short-term predictions.
\end{itemize}

\section{Method}
\label{sec:method}
\subsection{Multi-Scenario Multi-Physiological Signal Datasets}
\label{ssec:method-dataset}
Datasets: We use three datasets with simultaneous ECG, PPG, and continuous BP recordings. The first is the Multi-states dataset collected in a laboratory, the second is the Intraoperative dataset from a local hospital, and the third is a subset of MIMIC-III waveforms~\cite{johnson2016mimic}. Each dataset includes 20 subjects.

1) Multi-states dataset: Five distinct interventions were utilized to enhance BP variations, including lying, sitting, deepbreathing, playing games, and hanggrip. Throughout the session, ECG, PPG, and BP signals were simultaneously captured at a sampling rate of 1000 Hz.

2) Intraoperative dataset: Intraoperative clinical data, including ECG, PPG, and invasive BP signals, were collected from the hospital at a sampling rate of 250 Hz.

3) MIMIC-III dataset: Subset of MIMIC-III includes ECG, PPG and BP signals recorded from ICU patients at a sampling rate of 125Hz.

Preprocessing: ECG and PPG signals are preprocessed to remove noise and artifacts using techniques such as trend removal and outlier exclusion. Band-pass filtering removes baseline drift and high-frequency noises from ECG signals, with 5 Hz and 40 Hz as low and high cutoff frequencies respectively. PPG signals are low-pass filtered at 10 Hz cutoff to reduce noise.

Feature extraction: A total of 38 features are extracted from the ECG and PPG signals. Time-domain features include pulse arrival time, its variation, heart rate, and morphological aspects like peaks, valleys, and peak-valley differences of PPG. ECG and PPG waveform features include mean, absolute sum, variance, square, and cross-correlation. Nonlinear features include fuzzy entropy. This multi-domain feature set comprehensively captures BP dynamics from the ECG and PPG data.

\subsection{Time-series Forecasting Model for BP}
\label{ssec:tsmodel-bp}
To efficiently and accurately predict BP dynamics, we review state-of-the-art (SOTA) time series forecasting models, including Transformer~\cite{vaswani2017attention}, Informer~\cite{zhou2021informer}, Autoformer~\cite{wu2021autoformer}, Pyraformer~\cite{liu2021pyraformer}, Crossformer~\cite{zhang2022crossformer}, and TimesNet~\cite{wu2022timesnet}. TimesNet leverages 2D-variations of 1D time series data using convolutional neural networks for better representation learning. To further enhance BP forecasting, we propose an improved architecture based on Inception, which efficiently captures data periodicity, making it ideal for physiological signals like ECG, PPG, and BP that follow cardiac cycles.

We propose the Time-series Attention-based Blood Pressure Forecast Network (TABNet), specifically designed for BP forecasting. TABNet integrates the Time-series Attention-based Block (TABBlock) module in Section \ref{ssec:tabblock}, reducing parameters and enhancing performance compared to the standard TimesNet. As shown in Figure \ref{fig:totalframe}, TABNet embeds preprocessed multi-domain ECG and PPG features into the TABBlock, followed by decoding and de-normalization to generate personalized BP predictions.

\subsection{TABBlock}
\label{ssec:tabblock}
The physiological mechanism of blood pressure regulation varies among individuals. To predict BP in time series, the proposed TABBlock incorporates a parameter-free attention mechanism into the Inception module~\cite{szegedy2015going, wan2024swift}, allowing the model to focus on distinct features like arterial stiffness, peripheral resistance, and cardiac output. This enables efficient individualized modeling while improving performance. Given the extracted ECG, PPG features, and BP time series $\mathbf{X}_{\text{FTS}}\in \mathbb{R}^{L\times M}$, where $L$ is the number of heartbeat cycles and $M$ is the number of extracted features, we first obtain feature representations via an embedding layer: $\mathbf{X}_{\text{FTS}}=\operatorname{Embed}(\mathbf{X}_{\text{FTS}})$. For the $l$-th layer, the input is $\mathbf{X}_{\text{FTS}}^{l-1}\in \mathbb{R}^{L\times d}$, and the overall process is:
\begin{equation}
\label{equ1}
\mathbf{X}_{\text{FTS}_{\text{1D}}}=\sum_{l=1}^{k}(\mathbf{X}_{\text{FTS}_{\text{1D}}}^{l-1} + \operatorname{TABBlock}(\mathbf{X}_{\text{FTS}_{\text{1D}}}^{l-1}) )
\end{equation}
The residual connection can make the signal transmitted in the deep network without attenuation or distortion, which is conducive to training the network.
\begin{figure}[!t]
\centering
\includegraphics[width=\linewidth]{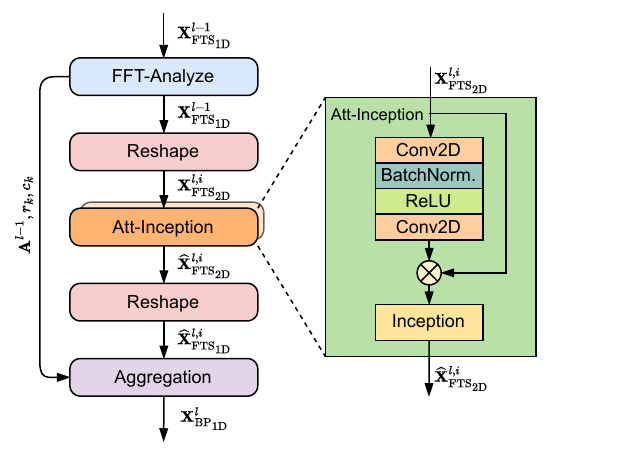}
\caption{The architecture of the proposed TABBlock denotes attention-based inception module.}
\label{fig:model_arch}
\end{figure}
TABBlock contains two successive parts: capturing 2D variations of BP time series and adaptively aggregating representations.







Firstly, we analyze the periodicity introduced by TimesNet \cite{wu2022timesnet}, by applying Fast Fourier Transform (FFT) to this 1D input, we can get $\mathbf{A}^{l-1}$, $r_i$ and $c_i$, which represents the amplitudes of all frequencies calculated, the top-$k$ important frequencies, and length of each period, while $c_{i} = \frac{T}{r_{i}}$, as well as rows and columns of the following 2D tensor:
\begin{equation}
\mathbf{A}^{l-1} = \operatorname{Amp_{avg}}\big(\operatorname{FFT}(\mathbf{X}_{\text{FTS}_\text{1D}}^{l-1})\big)
\end{equation}
\begin{equation}
r_{k} = \mathop{\arg\mathrm{Topk}}_{r\in{1,\cdots,\frac{T}{2}}}\left(\mathbf{A}^{l-1}\right)
\end{equation}
The above process can be simplified as:
\begin{equation}
\label{equ2}
\mathbf{A}^{l-1},{r_{k}},{c_{k}}=\operatorname{FFT_{\text{cal}}}(\mathbf{X}_{\text{FTS}_\text{1D}}^{l-1})
\end{equation}
This approach effectively extracts critical periodic characteristics from time-series data, crucial for the following analysis and forecasting temporal patterns.

Based on the significant frequencies $r_i$ and length we get $c_i$ from $\operatorname{FFT_{\text{cal}}}(\cdot)$, we reshape the 1D time series into 2D tensors by this $\operatorname{Reshape}(\cdot)$ operation, the shift from 1D to 2D enables the use of convolutional layers for extracting spatial features and identifying periodic trends in time-series data, enhancing the model's predictive accuracy:
\begin{equation}
\label{equ3}
\mathbf{X}_{\text{FTS}_{\text{2D}}}^{l,i}=\operatorname{Reshape}_{c_i,r_i}(\mathbf{X}_{\text{FTS}_\text{1D}}^{l-1}), i\in{1,\cdots,k}
\end{equation}
Then, we apply the improved Inception module incorporated with an attention mechanism in TABBlock to enhance feature learning. Unlike TimesNet which processes all features uniformly, the added attention layer enables focusing on informative components for more effective representation learning. Meanwhile, reducing model dimensionality improves efficiency. The attention-based lightweight TABBlock architecture improves prediction performance while maintaining high computational efficiency.
\begin{equation}
\label{equ4}
\mathbf{Z}_{\text{2D}}^{att}=\operatorname{Conv2D}(\operatorname{ReLU}(\operatorname{Conv2D}(\mathbf{X}_{\text{FTS}_{\text{2D}}}^{l,i})))
\end{equation}
\begin{equation}
\label{equ5}
\widehat{\mathbf{X}}_{\text{FTS}_\text{2D}}^{l,i} = \mathbf{X}_{\text{FTS}_{\text{2D}}}^{l,i} \otimes \mathbf{Z}_{\text{2D}}^{att}
\end{equation}
$\operatorname{Conv2D}(\cdot)$ module contains a $3 \times 3$ convolution layer, which enlarges the receptive field to extract more global feature information. The attention feature map $\mathbf{Z}_{\text{2D}}^{att}$ enhances feature learning by adaptively regulating the importance of input features.
\begin{equation}
\label{equ6}
\widehat{\mathbf{X}}_{\text{FTS}_\text{2D}}^{l,i}=\operatorname{AttInception}(\mathbf{X}_{\text{FTS}_\text{2D}}^{l,i}), i\in{1,\cdots,k}
\end{equation}
Where the $\operatorname{AttInception}(\cdot)$ module contains a lightweight attention layer to focus on informative features as shown in Figure \ref{fig:model_arch}. Finally, we transform the 2D features back to 1D.
\begin{equation}
\label{equ7}
\widehat{\mathbf{X}}_{\text{FTS}_\text{1D}}^{l,i} =\text{Reshape}_{1,(c_i\times r_i)}(\widehat{\mathbf{X}}_{\text{FTS}_\text{2D}}^{l,i}), i\in{1,\cdots,k}
\end{equation}
Finally, we also use $\operatorname{Aggregation}(\cdot)$ module, which is the adaptive aggregation method used in TimesNet to first truncate $\widehat{\mathbf{X}}_{\text{FTS}_\text{1D}}^{l,i}$, and then fuse $k$ truncated 1D output features $\widehat{\mathbf{X}}_{\text{FTS}_\text{1D}}^{l,i}$ based on the previously calculated frequency amplitude $\mathbf{A}^{l-1}$ corresponding to different periods, and then pass them to the next layer.
\begin{equation}
\label{equ8}
\mathbf{X}_{\text{BP}_\text{1D}}^{l} = \operatorname{Aggregation}({\mathbf{A}}^{l-1}\times \widehat{\mathbf{X}}_{\text{FTS}_\text{1D}}^{l,i})
\end{equation}
By modeling the time series of BP in both time and non-linear domains, TABBlock can effectively learn multi-scale temporal patterns, thereby improving prediction performance.

\begin{table*}[!t]
\centering
\caption{\textbf{The comparison results of the average MAE and SD of SBP predicted by our proposed personalized model TABNet on different scenarios corresponding three datasets after fixing the input length to 30 heartbeat cycles.}}
\begin{adjustbox}{width=0.85\linewidth,center}
\begin{threeparttable}
\begin{tabular}{ccccccccccc}
    \toprule 
      \multirow{3}{*}{\makecell[c]{Training Data\\(Heartbeat cycles)}} & \multirow{3}{*}{\makecell[c]{Forecast Length\\(Heartbeat cycles)}} & \multicolumn{2}{c}{Multi-states Data} & \multicolumn{2}{c}{Intraoperative Data} & \multicolumn{2}{c}{MIMIC-III Data}\\
      \cline{3-8}
      & & \makecell[c]{MAE\\(mmHg)} & \makecell[c]{SD\\(mmHg)} & \makecell[c]{MAE\\(mmHg)} & \makecell[c]{SD\\(mmHg)} & \makecell[c]{MAE\\(mmHg)} & \makecell[c]{SD\\(mmHg)} \\
      \hline
      \multirow{3}{*}{60} & 5 & 5.06 & 6.67 & 7.72 & 7.06 & 3.02 & 2.61 \\
      & 10 & 5.67 & 6.88 & 7.86 & 7.16 & 3.25 & 2.79 \\
      & 20 & 6.66 & 6.35 & 8.19 & 7.29 & 3.37 & 2.91 \\ \hline
      
      \multirow{3}{*}{180} & 5 & 4.77 & 5.41 & 7.26 & 6.83 & 2.92 & 2.54 \\
      & 10 & 5.15 & 5.42 & 7.29 & 6.77 & 3.13 & 2.72 \\
      & 20 & 5.55 & 4.97 & 7.77 & 7.01 & 3.31 & 2.86 \\ \hline
      
      \multirow{3}{*}{300} & 5 & 3.66 & 3.66 & 7.14 & 6.86 & 2.93 & 2.52 \\
      & 10 & 4.19& 4.04 & 7.36 & 7.01 & 3.10 & 2.68 \\
      & 20 & 4.67 & 3.78 & 7.77 & 7.23 & 3.32 & 2.85 \\ \hline

      \multirow{3}{*}{420} & 5 & 3.28 & 3.42 & 7.49 & 6.87 & 2.87 & 2.55 \\
      & 10 & 3.54 & 3.26 & 7.78 & 7.01 & 3.09 & 2.75 \\
      & 20 & 4.36 & 2.79 & 8.20 & 7.32 & 3.08 & 2.70 \\ 
      
     \bottomrule
\end{tabular}
\end{threeparttable}
\label{table:all_results}
\end{adjustbox}
\end{table*}

\section{Experiment}
\label{sec:experiment}
\subsection{Experimental Details}
\label{ssec:exp-details}
As for implementation details, the train, validation and test splits are set to 7:1:2 to enable testing with more data. The encoder input dimension is set to 39 (containing input signal features and BP values). All personalized models are trained for 10 epochs with a batch size of 4 and employ mean square error as loss function. The encoder input size is 39 to match the number of input features and BP. Adam optimizer is used for optimization with a learning rate of 1e-4. The $top_k$ hyperparameter is set to 5 to select the most dominant periodicities based on the FFT frequency distributions.
\subsection{Metrics and Results}
\label{ssec:results}
To analyze the generalization capability of TABNet across environments, we calculate the average MAE and SD for 20 subjects (60 in total) from each of the three distinct datasets. This allows us to examine whether the model meets AAMI standard consistently in diverse clinical scenarios.

For personalized models, we investigate the amount of individual patient data needed for effective modeling by varying the training sequence length from 60 to 420 heartbeat cycles. The input length is fixed at 30 cycles, and the forecasting sequence lengths are set to 5, 10, and 20 cycles to assess both short- and long-term capabilities. As shown in Table \ref{table:all_results}, longer training sequences reduce prediction errors, with MAE decreasing by up to 2.5 mmHg when using 420 cycles. This indicates that more data helps capture richer features, avoiding overfitting and improving BP forecasting. Additionally, longer prediction sequences show higher errors, consistent with time series forecasting principles. The best performance occurs when forecasting 5 heartbeat cycles with 420 training cycles, reaching 2.87 mmHg MAE and 2.55 mmHg SD on the MIMIC-III dataset. For the generalized model, the input length is fixed at 30 cycles. The experiments used data from 15 subjects for training and 5 for testing, with no overlap between sets.

For the intraoperative dataset, while the model meets AAMI standards, its performance is lower compared to the other datasets due to significant BP fluctuations, making it challenging for TABNet to capture stable feature representations for accurate short-term BP predictions.

Overall, these results validate that using individualized models with relatively less training data can achieve accurate real-time BP monitoring to meet clinical needs.

\subsection{Ablation Study}
In our comprehensive ablation study, we evaluate the performance of our TABNet model against a suite of SOTA time-series models including TimesNet~\cite{wu2022timesnet}, Transformer~\cite{zeng2023transformers}, Autoformer~\cite{wu2021autoformer}, Pyraformer~\cite{liu2021pyraformer}, and Crossformer~\cite{zhang2022crossformer}, these models are adapted and assessed for this specific task of personalized blood pressure forecasting. Each model, including our TABNet, is tested using their respective standard configurations across three datasets. The training data comprises 420 cycles with an output length of 20 cycles, ensuring a consistent and fair basis for comparison.

In the comparative analysis, as detailed in Table \ref{table:ablation}, we use grid search method~\cite{bao2006fast} to find the best hyperparameters and corresponding best performance for each model, our TABNet notably excells, surpassing all the SOTA models adapted for the BP task, both in terms of MAE and SD. This performance reinforces the effectiveness of TABNet’s architecture, especially its attention-based mechanism, in this specialized predictive context. The results not only demonstrate TABNet's superior capabilities but also establish its significant position in advanced time series modeling for personalized BP prediction.
\begin{table}[t]
\begin{center}
\caption{Ablation study of our proposed model. The best performance are shown as MAE $\pm$ SD in \textbf{bold black}.}
\begin{adjustbox}{width=\linewidth,center}
\begin{tabular}{lccc}   
\toprule
\multirow{2}{*}{Model} & \multirow{2}{*}{\makecell[c]{Multi-states\\Data}} & \multirow{2}{*}{\makecell[c]{Intraperative\\Data}} & \multirow{2}{*}{\makecell[c]{MIMIC-III\\Data}} \\ \\
\midrule
    Autoformer~\cite{wu2021autoformer} & $4.67 \pm 3.27$ & $13.83 \pm 12.42 $ & $6.35 \pm 5.24$ \\
    Crossformer~\cite{zhang2022crossformer} & $3.49 \pm 2.44$ & $11.24 \pm 10.09$ & $4.52 \pm 4.21$ \\ 
    Transformer~\cite{vaswani2017attention} & $3.98 \pm 2.82$ & $8.71 \pm 7.82$ & $4.02 \pm 3.33$ \\ 
    Pyraformer~\cite{liu2021pyraformer} & $2.91 \pm 2.04$ & $8.81 \pm 7.87$  & $4.89 \pm 4.35$ \\ 
    TimesNet~\cite{wu2022timesnet} & $2.40 \pm 1.68$ & $7.73 \pm 6.94$ & $3.55 \pm 2.96$ \\ 
    \textbf{TABNet (Ours)} & $\textbf{2.02} \pm \textbf{1.43}$ & $\textbf{7.60} \pm \textbf{6.85}$ & $\textbf{3.11} \pm \textbf{2.78}$ \\
\bottomrule   
\end{tabular}   
\label{table:ablation}
\end{adjustbox}
\end{center}
\end{table}

\section{Conclusion}
\label{sec:conclusion}
In this paper, we propose the personalized TABNet model for continuous BP forecasting in critical care. TABNet employs attention-based time series modeling to capture physiological relationships within and across cardiac cycles. The 2D representation learning enhances periodic pattern modeling. TABNet demonstrates accurate short-term forecasting on diverse clinical datasets, enabled by the tailored architecture and lightweight design allowing efficient training with limited individual data. TABNet meets AAMI standards consistently, except on two tasks using highly fluctuating intraoperative data. The proposed personalized framework provides a practical tool for early abnormal BP detection and timely intervention for at-risk patients. This can facilitate continuous non-invasive BP monitoring to reduce mortality and improve prognosis in clinical care.
\section*{Acknowledgments}
Thanks to Min Tang, Zengding Liu and Fen Miao for the intraoperative data provided.

\bibliographystyle{IEEEtran}
\bibliography{IEEEabrv, myref}

\end{document}